\documentclass[sigconf]{acmart}
\AtBeginDocument{%
  }

\usepackage{graphicx}
\usepackage{xspace}

\newcommand{\bph}{\ensuremath{\textsc{BharatPotHole}}\xspace}
\newcommand{\rw}{\ensuremath{\textsc{iWatchRoad}}\xspace}
\copyrightyear{2025}
\acmYear{2025}
\setcopyright{cc}
\setcctype{by-nc-nd}
\acmConference[CODS 2025]{13th ACM IKDD International Conference on Data Science}{December 17--20, 2025}{Pune, India}
\acmBooktitle{13th ACM IKDD International Conference on Data Science (CODS 2025), December 17--20, 2025, Pune, India}
\acmDOI{10.1145/3799830.3799873}
\acmISBN{979-8-4007-2355-1/2025/12}
\begin{document}

\title{iWatchRoad: Scalable Detection and Geospatial Visualization of Potholes for Smart Cities}

\author{Rishi Raj Sahoo}
\authornote{Both authors contributed equally to this research.}
\email{rishiraj.sahoo@niser.ac.in}
\affiliation{%
  \institution{National Institute of Science Education and Research (NISER), An OCC of Homi Bhabha National Institute}
  \city{Bhubaneswar}
  \state{Odisha}
  \country{India}
}
\author{Surbhi Saswati Mohanty}
\authornotemark[1]
\authornote{This work was carried out while the author was an intern at NISER.}
\email{cen.22bced69@silicon.ac.in}
\affiliation{%
  \institution{Silicon University}
  \city{Bhubaneswar}
  \state{Odisha}
  \country{India}
}

\author{Subhankar Mishra}
\email{smishra@niser.ac.in}
\affiliation{%
  \institution{National Institute of Science Education and Research (NISER), An OCC of Homi Bhabha National Institute}
  \city{Bhubaneswar}
  \state{Odisha}
  \country{India}
}

\renewcommand{\shortauthors}{Sahoo et al.}

\begin{abstract}
Potholes on the roads are a serious hazard and maintenance burden. This poses a significant threat to road safety and vehicle longevity, especially on the diverse and under-maintained roads of India. In this paper, we present a complete end-to-end system called \rw for automated pothole detection, Global Positioning System (GPS) tagging, and real time mapping using OpenStreetMap (OSM). We curated a large, self-annotated dataset of over 7,000 frames captured across various road types, lighting conditions, and weather scenarios unique to Indian environments, leveraging dashcam footage. This dataset is used to fine-tune, Ultralytics You Only Look Once (YOLO) model to perform real time pothole detection, while a custom Optical Character Recognition (OCR) module was employed to extract timestamps directly from video frames. The timestamps are synchronized with GPS logs to geotag each detected potholes accurately. The processed data includes the potholes' details and frames as metadata is stored in a database and visualized via a user friendly web interface using OSM. \rw not only improves detection accuracy under challenging conditions but also provides government compatible outputs for road assessment and maintenance planning through the metadata visible on the website. Our solution is cost effective, hardware efficient, and scalable, offering a practical tool for urban and rural road management in developing regions, making the system automated. \rw is available at \url{https://smlab.niser.ac.in/project/iwatchroad}
\end{abstract}

\ccsdesc[500]{Computing methodologies~Computer vision}
\ccsdesc[300]{Computing methodologies~Object detection}
\ccsdesc[300]{Information systems~Spatial-temporal systems}
\ccsdesc[100]{Applied computing~Transportation}

\keywords{
Pothole detection, YOLO, GPS tagging, OpenStreetMap, Computer vision, Road maintenance, Real-time detection, Indian roads}
\begin{teaserfigure}
  \includegraphics[width=\textwidth]{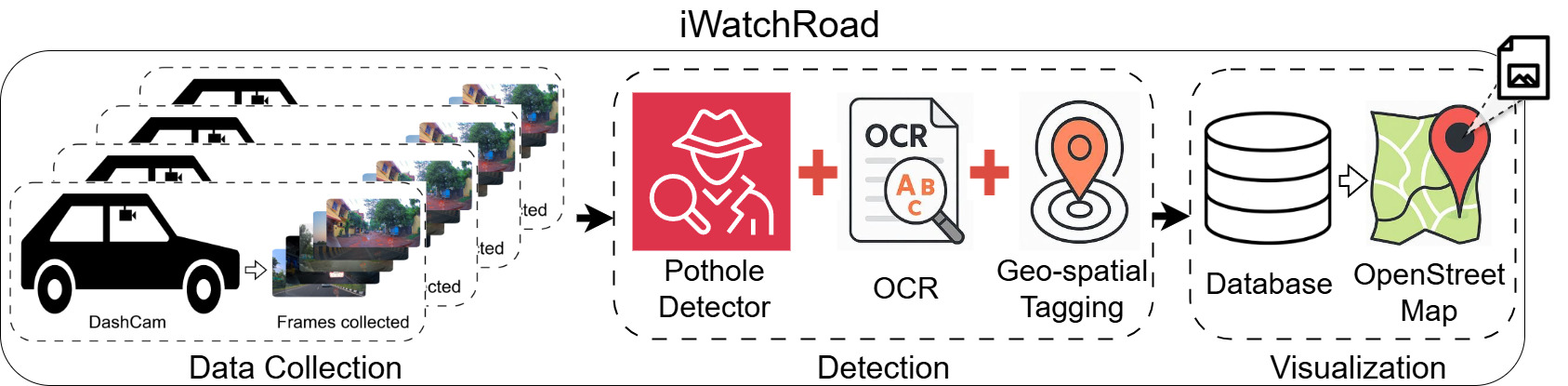}
  \caption{This simplified pipeline showcases the core stages of the \rw from dashcam based data collection, frame extraction, and pothole detection (using OCR and geospatial tagging), to visualization on a map based interface using OpenStreetMap.}
  \Description{}
  \label{fig:teaser} 
\end{teaserfigure}


\maketitle

\section{Introduction}
Maintaining road infrastructure is critical to ensure safe transportation and support economic development. Potholes are one of the most common and dangerous pavement problems. They increase the risk of accidents and spinal injuries caused due to sudden jolts. Automobiles suffer damage and reduced lifespan due to excessive braking to avoid potholes. In India, potholes pose significant safety hazards, with the Ministry of Road Transport and Highways (MoRTH) reporting 2,140 and 1,471 road accident deaths due to potholes in 2019 and 2020, respectively \cite{11031494}. Early detection and repair of potholes is essential for road safety and infrastructure maintenance \cite{Safyari2024}. Although Municipal corporations of cities such as Pimpri Chinchwad (PCMC, Maharashtra) and some cities of Gujarat still rely on manual inspection or citizen reports, which are slow and labor intensive. There is a need for automated, real time solutions that can streamline road condition monitoring and facilitate timely repair. 

Recent advances in computer vision and deep learning models have shown promising results in automated pothole detection. Several studies have trained these models on custom datasets to detect potholes from street view images. Specifically, most of these works either rely on limited datasets such as 1,500 images or use foreign datasets that do not reflect the full variability of Indian roads, especially rural, poorly lit, or weather affected conditions. While some systems integrate GPS tagging and mapping, their focus is often limited to urban roads or forest paths without comprehensive metadata or detailed visual information. Furthermore, prior works rarely include negative samples (e.g., manhole covers, shadows) to reduce false positives or fail to handle timestamp synchronization effectively.

Earlier systems lacked geospatial integration or struggled with Indian road diversity. Our approach introduces a comprehensive solution that begins with dashcam footage and ends with real time visualization on a web platform as shown in Figure \ref{fig:teaser}. Dashcams have become increasingly popular among the public, with statistics showing that around 7.2 million units were purchased by Indians in FY 2024. This widespread adoption enables our system to operate automatically, allowing users to contribute without any manual effort. We address the absence of geotagging in prior work by synchronizing Optical Character Recognition
(OCR) \cite{easyocr} extracted timestamps with external GPS logs, ensuring every detected pothole is accurately located. Furthermore, we tackle the challenge of domain mismatch and low light failures by building and fine-tuning our YOLO \cite{redmon2016you} model on a diverse, self-curated Indian dataset featuring varied road types, weather, and lighting conditions: a major limitation in previous global datasets. By incorporating negative samples (non-pothole road frames), our model is also trained to minimize false positives, such as shadows or manhole covers. Moreover, it uploads the metadata rich results (timestamp, GPS, severity, contractor information) to a live, OpenStreetMap (OSM) \cite{openstreetmap} based interface. This ensures transparent, scalable, and actionable insights for both government authorities and the public, going beyond just detecting potholes. Further, when potholes are repaired the database is updated accordingly.

\subsection*{Contribution}
\begin{enumerate}
    \item \bph: We introduce a large self-annotated dataset of Indian roads, capturing diverse road types, weather, and lighting conditions to reflect real world driving scenarios.
    \item Custom trained detection model: We released a pothole detection, deep learning based YOLO model, fine tuned on \bph, achieving strong performance under challenging conditions.
    \item OCR based GPS synchronization: We synchronized dashcam video frames with an external GPS track by extracting frame timestamps via OCR, enabling precise geotagging of potholes.
    \item Integrated mapping platform: We developed a web interface using OpenStreetMap that displays detected potholes with frame and metadata, creating a live pothole map.
    \item Metadata Information: The \rw platform provides metadata, including assessment of pothole condition and temporal information, through the web interface.
\end{enumerate}

\section{Related Work}

This section highlights the existing work on pothole detection and its tagging.

\subsection*{Vision-based Pothole Detection}
Computer vision approaches for pothole detection have evolved significantly in recent years. Traditional image processing methods such as thresholding and segmenting have been used, but recent work focuses on deep neural networks \cite{Ma_2022}. In particular, object detection CNNs like Faster R-CNN, SSD, and YOLO have shown high accuracy and speed for real-time pothole detection. This paper \cite{9112424} curated a 1500 image dataset of Indian roads and trained YOLOv3, YOLOv2, and YOLOv3-tiny models to detect potholes. However, their setup is not easily scalable, as the angle of the image captured through mobile phones is not consistent across samples. Furthermore, the dataset is not publicly available, and the authors have not incorporated any geotagging of the collected data. Similarly, \cite{10.1007/s42979-024-02887-1} compared YOLOv5/6/7 on potholes from various roads in India and found YOLOv7 attained the highest precision. These works demonstrate that modern CNN detectors, such as YOLO are highly effective in detecting road defects. Stereoscopic and 3D methods are other vision approaches. Stereo cameras and LiDAR can reconstruct the geometry of the road to locate potholes, but these setups are costly and complex \cite{8577119}. Mobile sensors (accelerometers or ultrasonic) have been used for pothole signaling, yet they lack spatial detail.

\begin{figure*}[h]
  \centering
  \includegraphics[width=\linewidth]{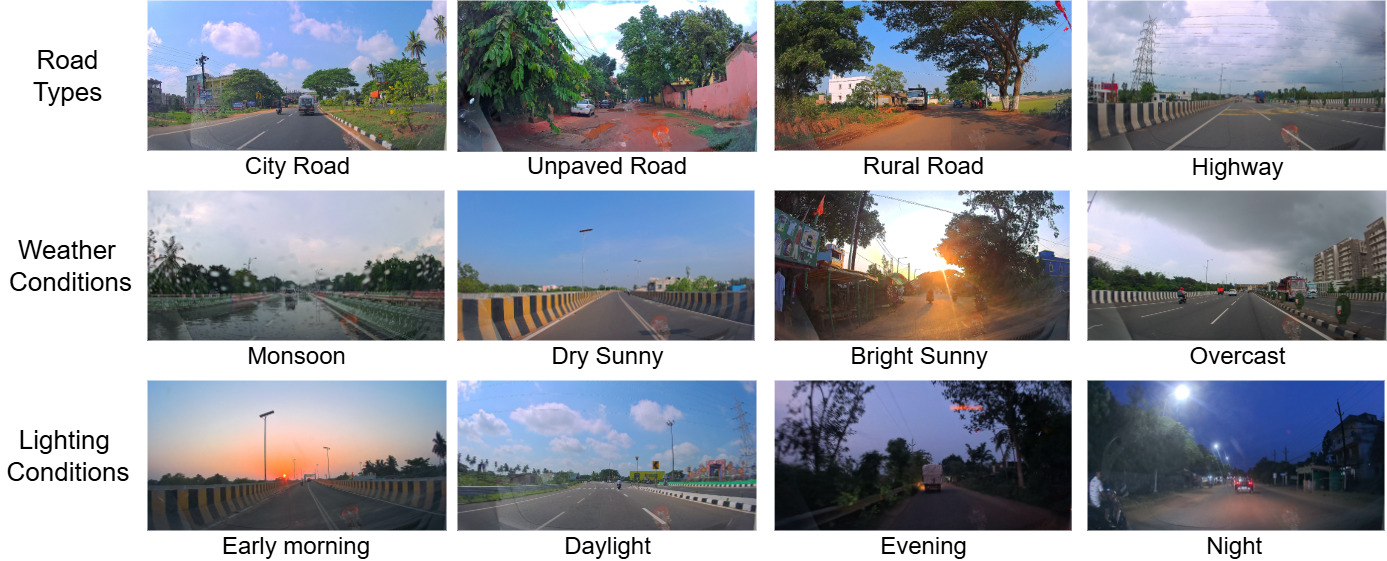}
  \caption{A comprehensive dataset covering diverse road types, weather, and lighting conditions to fine-tune the model for improved robustness and real-world performance.}
  \Description{}
  \label{fig:dtype}
\end{figure*}

\subsection*{Mapping and Geotagging Potholes}
GPS based location tagging has been extensively used in road infrastructure monitoring. Studies have shown that GPS coordinates can be effectively extracted from dashcam footage and integrated with pothole detection systems \cite{8901626}. The paper \cite{hoseini2023pothole} has developed a deep learning based pothole detection and tracking system for forest roads using dashcam footage. They used object detection and tracking to geolocate potholes and maps for maintenance planning. safeDrive \cite{11042541} is a well organized system that uses YOLOv7 on foreign dataset and cloud based tools for detection and mapping. These works lack diverse Indian data and information about the potholes.

\subsection*{Other Related Work}
There are several works that are based on vehicle based pothole alert systems. Smart Vehicle mounted frameworks combine deep learning with database logging to warn drivers and update authorities in real time \cite{Lincy}. Another related line of research uses thermal or stereo imaging for the detection of potholes in challenging conditions \cite{Yebes_2021}. These methods improve robustness, while they still require geospatial tagging.

\section{\bph: Curation, Diversity and Annotation}
A comprehensive pothole detection dataset was curated through the systematic collection of dashcam footage from various Indian roads. Data was collected using dashcam-equipped vehicles over a three-month span (May-July), covering a wide range of Indian road and weather conditions. The dataset encompasses multiple road categories, including unpaved rural pathways, urban arterial roads, village thoroughfares, and national highways. Environmental diversity was ensured by capturing footage under varied weather conditions, from monsoon rains to dry sunny days. Temporal variation was incorporated through data collection across different lighting scenarios: bright daylight, dawn/dusk conditions, and nighttime with street lighting. Figure \ref{fig:dtype} illustrates representative samples demonstrating the diverse characteristics of the collected dataset. Such diversity is essential for developing robust pothole detection systems capable of generalizing across real-world deployment scenarios.

Manual annotation was performed on all frames containing visible pothole instances, with precise bounding box annotation using Roboflow \cite{dwyer2025roboflow}. Subsequently, annotations were converted to YOLO format to facilitate model training compatibility. The resulting dataset comprises more than 7,000 annotated frames, all acquired through our dashcam recordings, representing the most extensive pothole detection dataset specifically tailored for Indian road conditions. Using only our dashcam recordings ensures consistency and authenticity, making the collection particularly suitable for fine-tuning state-of-the-art detection models such as YOLOv8 \cite{yolov8_ultralytics} for Indian road applications.

Dataset partitioning follows established machine learning conventions with 70\% allocated for training, 20\% for validation, and 10\% for testing purposes. All images and corresponding labels maintain consistent formatting at 640×640 pixel resolution, adhering to standardized practices in object detection dataset preparation. The complete dataset is publicly available at 
\url{www.kaggle.com/datasets/surbhisaswatimohanty/bharatpothole}

\subsection{Pothole Data Perspective}

\begin{figure}[ht]
  \centering
  \includegraphics[width=\linewidth]{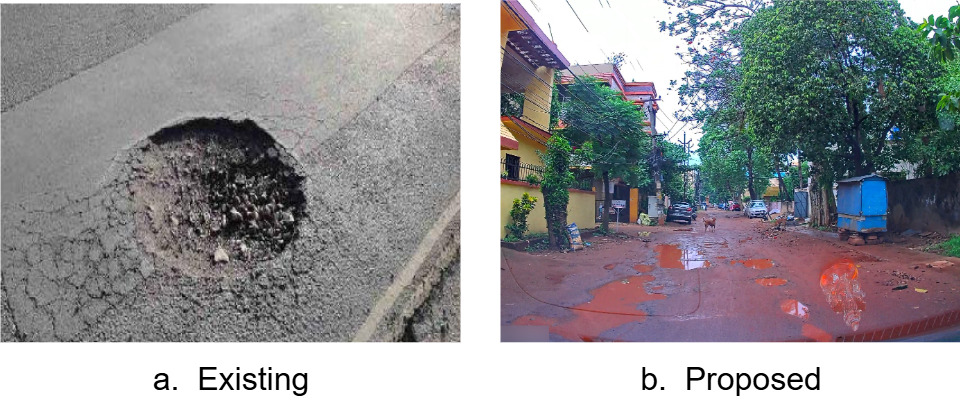}
  \caption{Comparison between traditional top-down pothole dataset (a), and the proposed forward facing dashcam view (b), which reflects real world driving scenarios and supports practical, continuous data collection for robust pothole detection.}
  \Description{}
  \label{fig:pv}
\end{figure}

The proposed dataset captures road imagery from the driver's perspective, providing a practical approach to continuous vehicle data acquisition, as demonstrated in Figure \ref{fig:pv}. This forward facing viewpoint corresponds directly to the input encountered by deployed pothole detection systems in real world driving scenarios. This approach contrasts with traditional pothole datasets that predominantly utilize top-down or close-up images captured from stationary positions directly above pothole instances. Although such elevated perspectives provide enhanced morphological detail, they lack practical relevance for automotive applications and present significant logistical challenges for continuous data collection.
Using dashcam footage streamlines data collection by eliminating requirements for specialized camera configurations or costly equipment installations. However, this perspective introduces inherent detection challenges: distant potholes appear as indistinct dark regions on the roadway, potentially leading to misclassifications of benign shadows or surface variations as pothole instances. The extensive scale and comprehensive annotation of the dataset enable robust model training to discriminate authentic pothole features from spurious visual artifacts.

\subsection{Environmental Variability and Robustness}
Pothole visibility demonstrates significant degradation under adverse environmental conditions, with instances readily detectable in optimal daylight becoming virtually imperceptible during nighttime. Precipitation further compounds detection complexity through surface reflections and altered shadow patterns. Models trained only on clear daylight images performed poorly in low light or wet conditions.

The environmental diversity incorporated within the dataset addresses these limitations by exposing the model to comprehensive operational conditions. This exposure enables the learning algorithm to develop discriminative features for authentic pothole detection while maintaining robustness against confounding factors such as shadows, reflections, and surface moisture. The inclusion of challenging environmental conditions significantly enhances overall system robustness across diverse deployment scenarios.
\subsection{Negative Sample Integration}

The dataset incorporates extensive negative samples that include road imagery in various conditions and surface types without pothole instances. These frames contain various characteristics of the road surface, including cracks, shadows, debris, petroleum stains, and utility access covers, none of which are annotated as pothole instances. The inclusion of negative samples is critical for preventing false positive detections that would otherwise compromise the system's reliability.
\begin{figure}[ht]
\centering
\includegraphics[width=\linewidth]{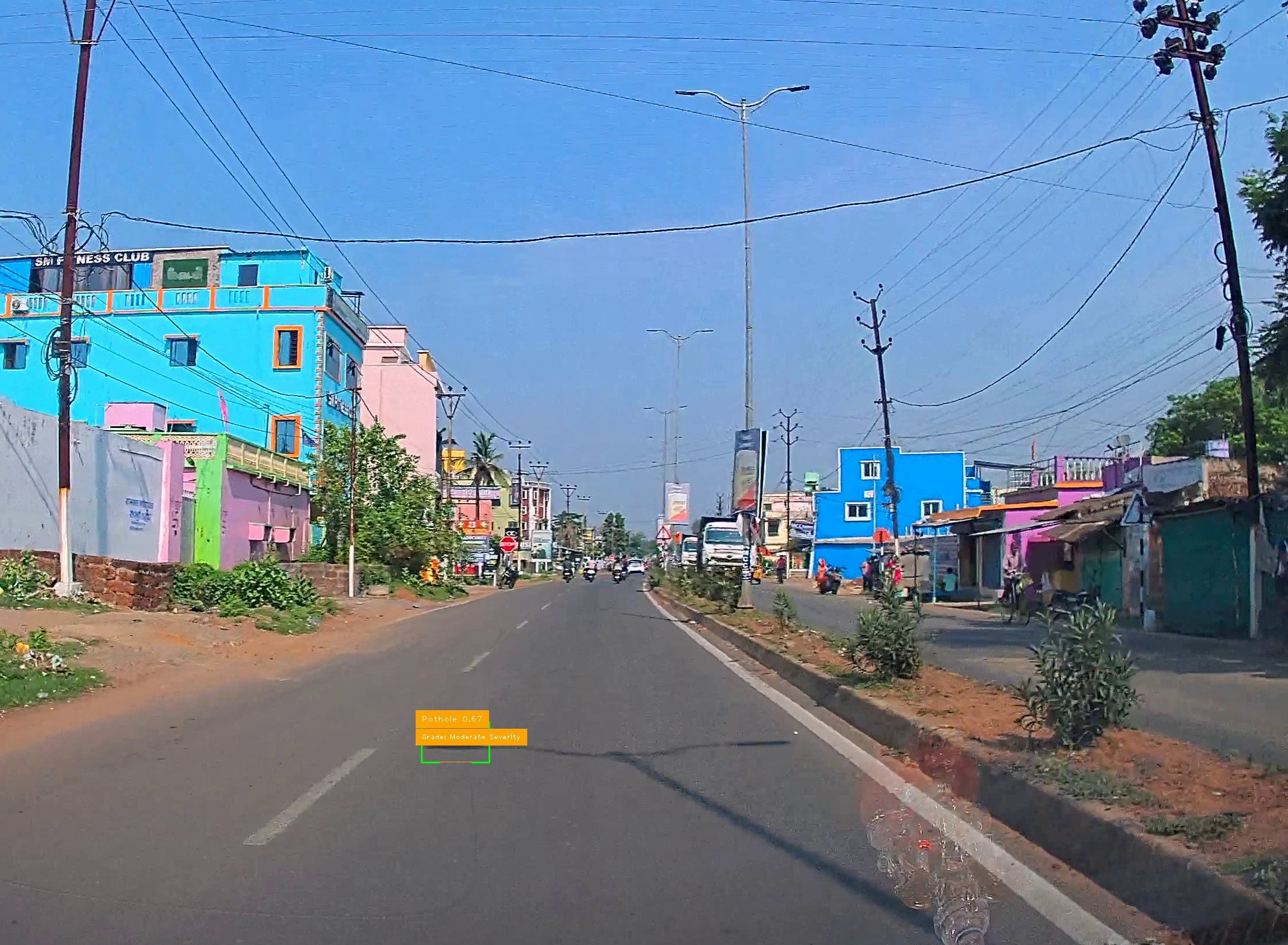}
\caption{Shadow misclassification as a pothole demonstrates the necessity of comprehensive negative sample training for robust feature discrimination.}
\Description{}
\label{fig:Lamp}
\end{figure}

Without adequate negative examples, trained models demonstrate a tendency to misclassify visually similar road features, such as utility covers or shadows, as pothole instances based solely on superficial appearance characteristics such as surface darkness. Figure \ref{fig:Lamp} illustrates this phenomenon when models lack sufficient negative sample exposure during training. The incorporation of these negative samples enables the learning algorithm to develop discriminative features based on authentic pothole characteristics, including irregular morphology, surface discontinuity, and edge definition, rather than relying on simplistic appearance cues such as dark surface patterns.

Experimental validation demonstrated that models trained exclusively on positive pothole instances exhibit elevated false positive rates. The balanced integration of positive pothole and negative road surface samples facilitates robust learning of pothole specific morphological features (irregular depressions with deteriorated boundaries) rather than generalized dark surface pattern recognition. This balanced approach significantly improves the specificity of the model and the overall detection performance in operational environments.

\begin{figure*}[!htbp]
  \centering
  \includegraphics[width=0.9\linewidth, height=5.5cm]{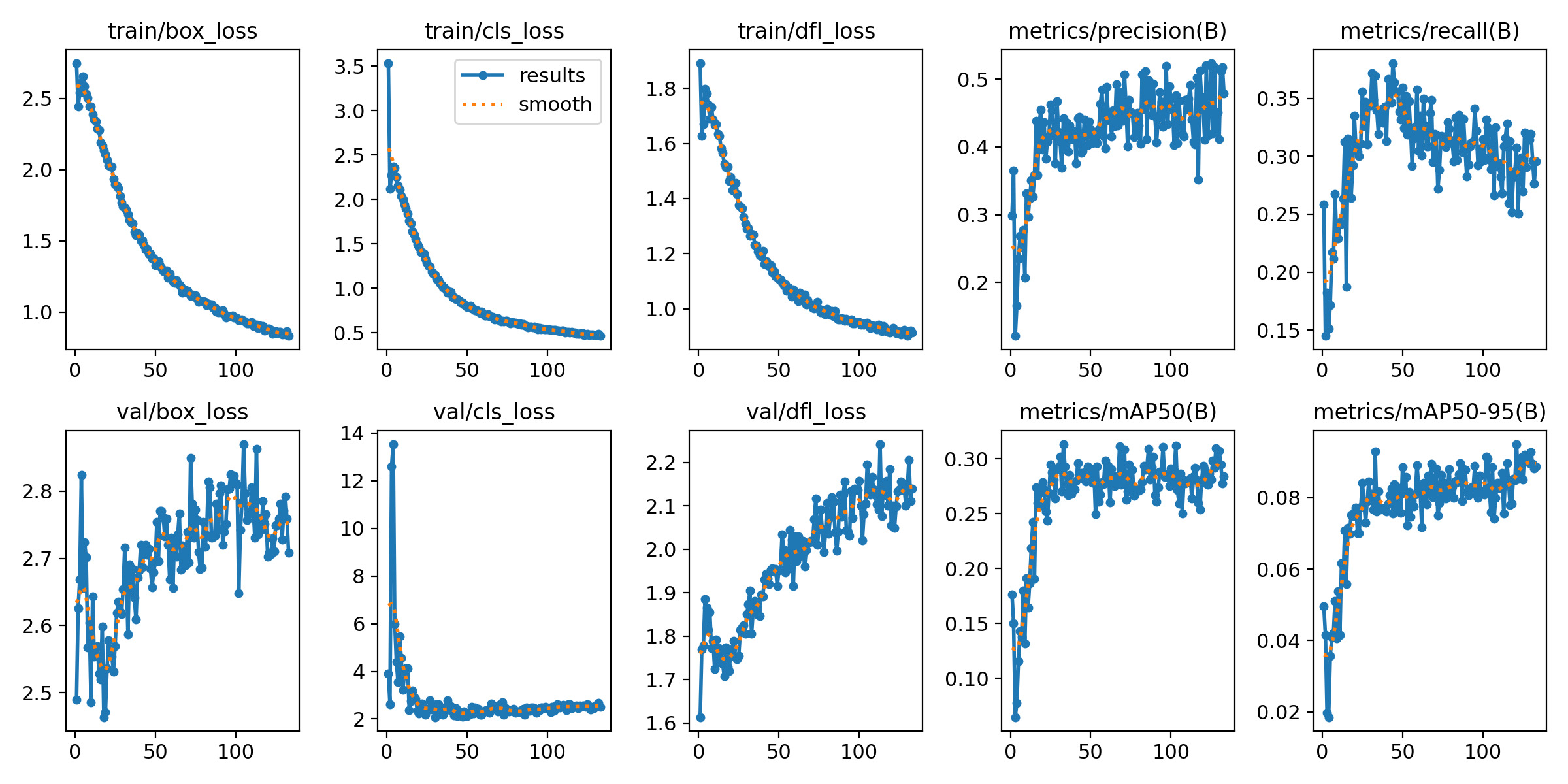}
  \caption{Model's performance on \bph-3K}
  \Description{}
  \label{fig:metric3k}
\end{figure*}
\begin{figure*}[!htbp]
  \centering
  \includegraphics[width=0.9\linewidth, height=5.5cm]{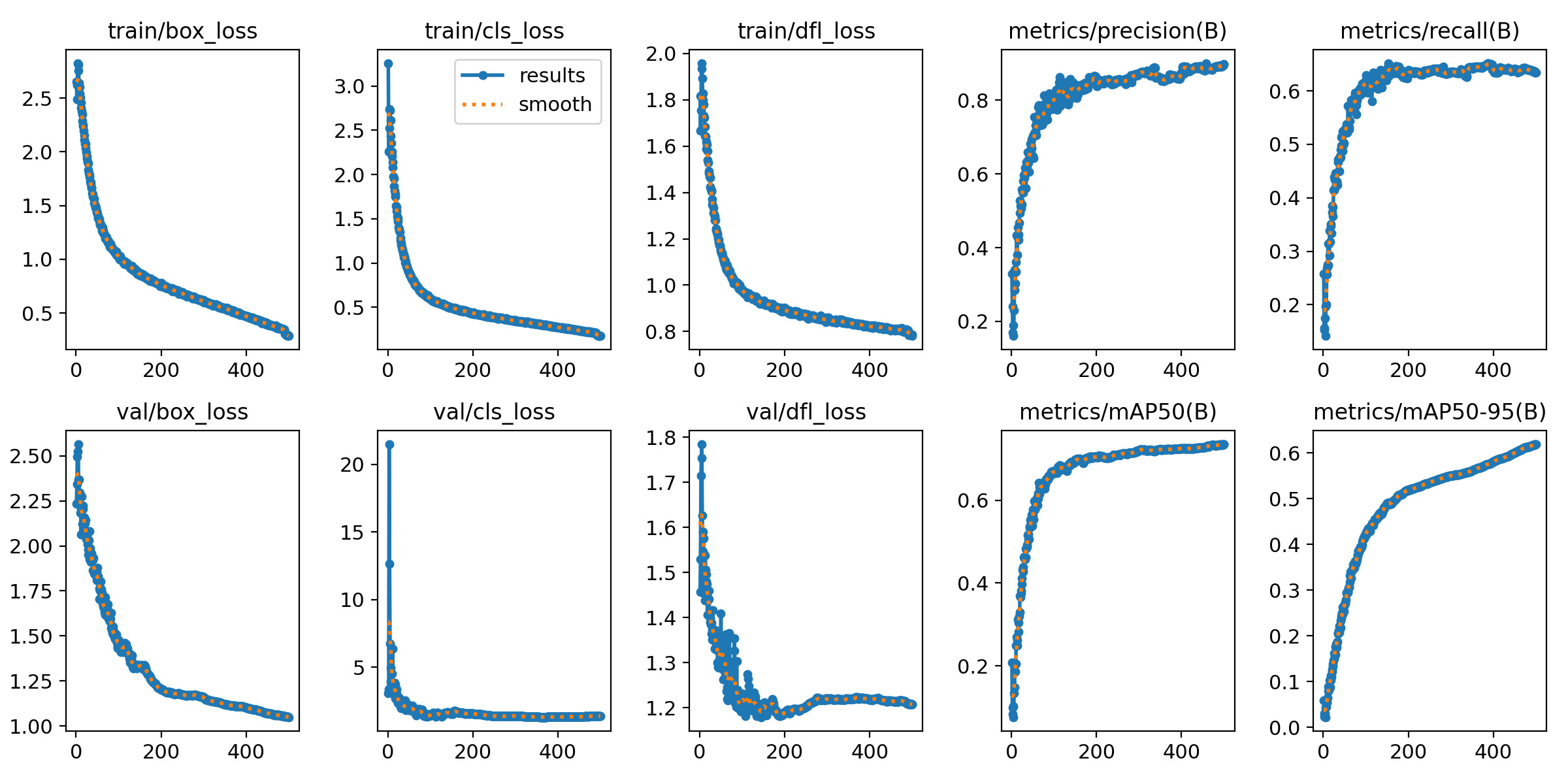}
  \caption{Model's performance on \bph-7K}
  \Description{}
  \label{fig:metric7k}
\end{figure*}
\section{ \rw }
This section outlines the architectural design of the \rw system. Indian roads present unique challenges for automated detection systems due to diverse surface materials, varying construction quality, and complex weather patterns. The architecture addresses these with a modular and scalable structure. 
\subsection{Data Collection}
 When users use dashcams, more data is collected without manual labor, which further adds to our server. This enhances the cost effectiveness of the data collection process. Personal identifiers such as license plate numbers and addresses are blurred before processing, ensuring anonymity.
\subsection{Detection}
The detection setup comprises three key modules:
\subsubsection{Custom trained YOLOv8 Detector:}
The model is fine-tuned using the \bph dataset, which includes thousands of annotated frames reflecting real time Indian road conditions, such as poorly lit, rain affected, or unpaved roads. The model's performance increased drastically with an increase in the dataset size from \bph-3k to \bph-7k, as shown in Figure \ref{fig:metric3k} and \ref{fig:metric7k} respectively. The fine-tuning process allows the model to distinguish potholes from similar looking artifacts such as shadows, manhole covers, or road markings. The output consists of bounding boxes around the detected potholes along with a confidence score, which is used to assess severity according to the size and shape.
\begin{figure*}[h]
  \centering
  \includegraphics[width=\linewidth]{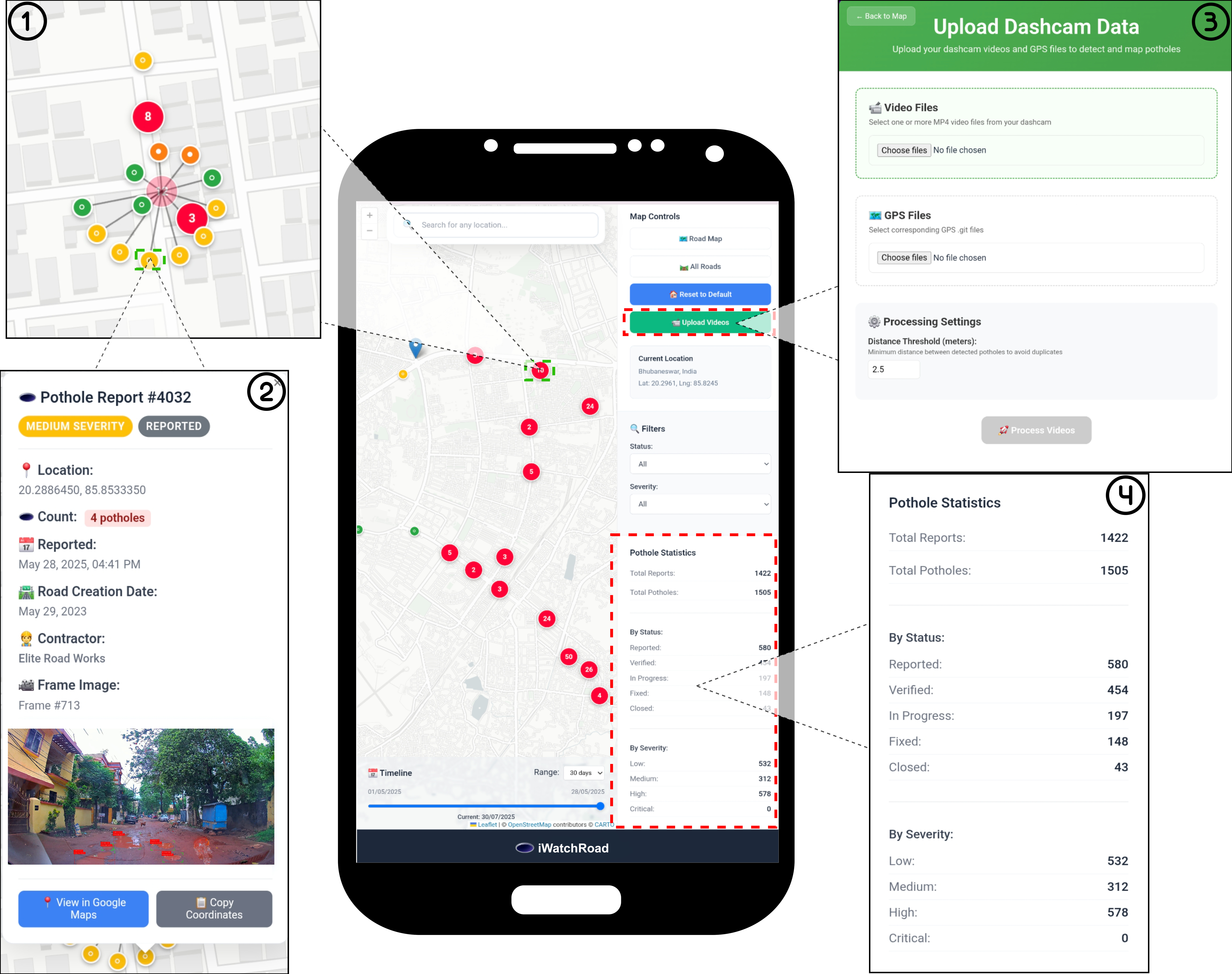}
  \caption{ \rw's Web Platform Interface: Interactive map view showing geotagged pothole reports with color coding based on severity,(1) shows a cluster of potholes, and when clicked, it shows (2) detailed metadata for each pothole, including timestamp, location, and frame image. (3) Upload interface where users can submit dashcam video and GPS files for automatic pothole detection and mapping. (4) Enumerates the pothole status.}
  \Description{}
  \label{fig:web}
\end{figure*}
\subsubsection{Optical Character Recognition (OCR):}
EasyOCR with regular expression (regex) parsing is used to extract and standardize embedded timestamps from video frames. Since dashcam overlays often vary in format and positioning, a custom regex parser is applied to reformat and clean the extracted strings into a consistent DD-MM-YYYY HH:MM:SS format. This ensures reliable alignment of each frame with the corresponding GPS data. OCR is implemented modularly, so it can be triggered only on selected frames to optimize performance and resource usage.
\subsubsection{Geo-spatial Tagging:}
GPS loggers continuously record the vehicle’s location and orientation. The timestamps in the logs are matched to the frame timestamps to geotag pothole detections.    

\subsection{Visualization}
This component includes a structured database and a web visualization layer:
\begin{figure*}[h]
  \centering
  \includegraphics[width=\linewidth]{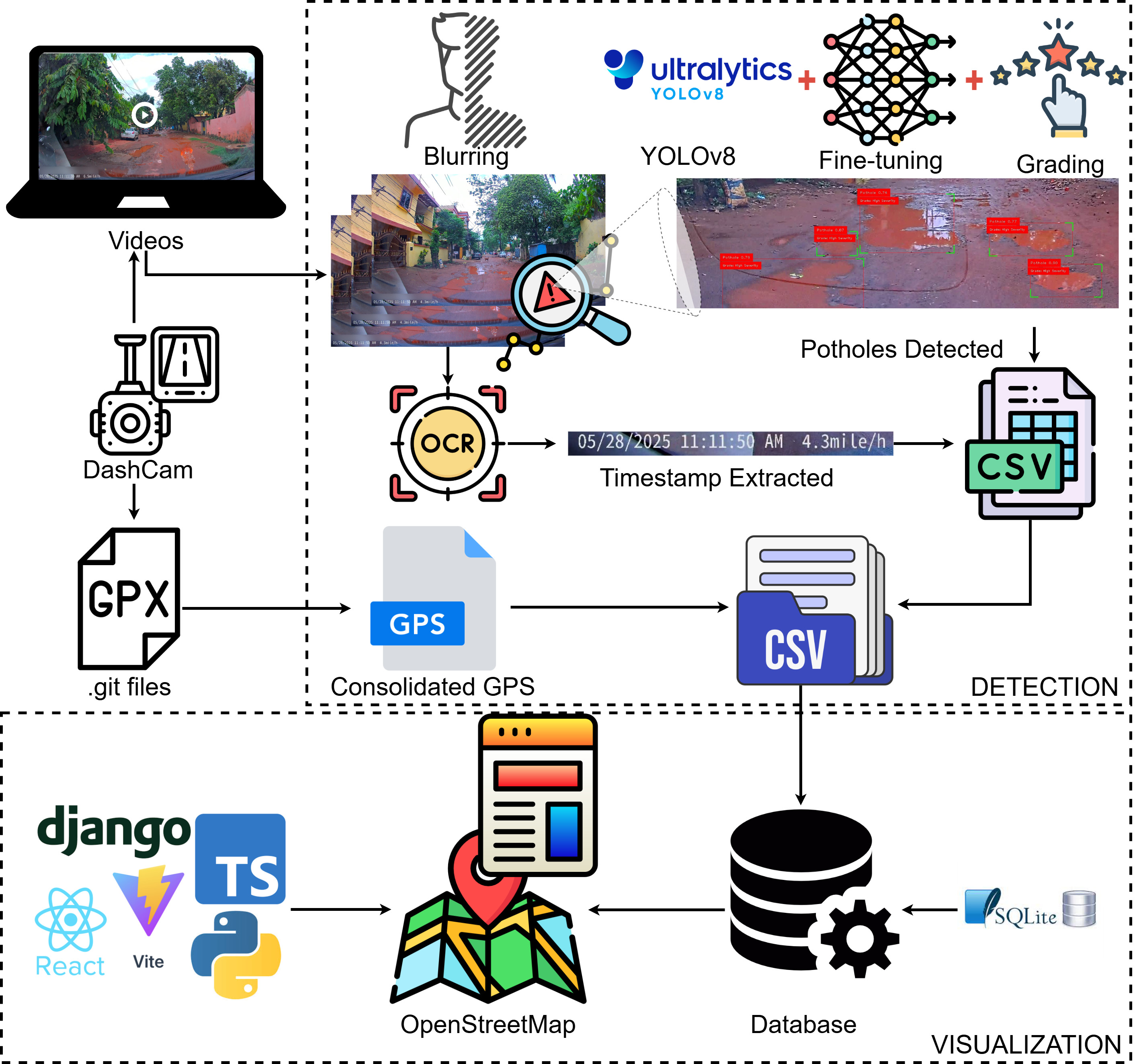}
  \caption{The diagram illustrates the end-to-end workflow of our pothole detection system. Videos captured via dashcams are converted into frames, in which the private information, like human faces and license plate numbers, is blurred. Frames are processed using a YOLOv8 model that has been fine-tuned for pothole detection. OCR extracts timestamps from frames, which are then matched with GPS data from .git files for accurate geotagging. The results are stored in CSV format and visualized on a web platform using OpenStreetMap, supported by a Django, React, and TypeScript stack, and backed by an SQLite database.}
  \Description{}
  \label{fig:roadwatch}
\end{figure*}
\subsubsection{Database}
A SQLite database stores metadata such as timestamps, coordinates, severity, and frame references.
\subsubsection{OpenStreetMap (OSM)}
The geotagged pothole records are visualized on a web application built with OpenStreetMap using Leaflet.js. Figure \ref{fig:web} illustrates the website's interface. Each pothole is marked by an icon on the map, highlighting the number of potholes at that location. The icon, when clicked, shows a pop up with metadata: a mini frame of the location, time of detection, severity of the pothole, road creation date, and contractor name. We host this on our server so users can zoom to any region of India and see the reported potholes. Our system also allows filtering by date or road type. This platform thus provides a live pothole map: for example, a cluster of markers indicates a badly damaged road needing urgent repair. Once the road is repaired and new dashcam data is uploaded, the server automatically removes the previous marker and mark it as fixed. The website is streaming on \url{https://smlab.niser.ac.in/project/iwatchroad}.

The pipeline is very simple; new data can be uploaded with the code simply by uploading the dashcam video and the external GPS source. It will detect, geotag, upload to the database, and start reflecting on the website. This is a modular pipeline, which can be updated with new interventions of technology in the future. The code is available at \url{https://github.com/smlab-niser/iwatchroad}

\section{Experiment}
The \rw web based pothole detection and reporting system was designed and validated through a comprehensive end-to-end implementation involving real world road data collection, AI based detection, database integration, and public-accessible web visualization. The experiment evaluates the system's ability to automatically detect potholes, geolocate them accurately, and present actionable information through an intuitive web platform accessible to both authorities and citizens.
\subsection{Experiment Setup}
The experimental validation was conducted on standard development hardware, demonstrating the system's accessibility and cost effectiveness. The implementation ran efficiently on multi-core Intel i5/i7 processors with 16 GB RAM and modest storage requirements, eliminating the need for specialized high-end infrastructure. The web application architecture comprised a Django backend (Python 3.11+) with React-TypeScript frontend and Vite build system, while mapping functionality utilized Leaflet.js through React-Leaflet integration. This technology stack ensures cross-platform compatibility and responsive web performance across devices, as illustrated in Figure \ref{fig:roadwatch}.

Data collection employed dashcam mounted vehicles with external GPS loggers to capture real world road conditions. Video streams were processed at 30 FPS with privacy protection through OCR based information detection and blurring. YOLOv8 was selected for real time detection capabilities and fine-tuned on region specific pothole datasets using RTX 3090 GPU acceleration. The detection pipeline generated bounding boxes, pothole counts, and severity assessments based on dimensional analysis.

The geotagging process synchronized OCR extracted timestamps from video overlays with GPS log entries to achieve precise spatial-temporal localization. All detection metadata, including coordinates, timestamps, severity classifications, and frame data was stored in a SQLite database optimized for web delivery. Base64 encoding of detection frames eliminated file system dependencies while enabling efficient frontend rendering and popup visualization on the web interface.

The experiment validated the system's cost effectiveness compared to expensive LiDAR or stereo camera solutions by utilizing consumer grade dashcams and GPS modules. Local processing capabilities on standard hardware eliminated cloud computing dependencies, while open-source technologies (Leaflet, Django, React, SQLite) minimized licensing costs. The resulting web platform demonstrates that effective infrastructure monitoring can be achieved through accessible, scalable web technologies that serve both automated data collection and public transparency objectives.
\section{Discussion and Analysis}

The advantages and challenges while implementing \rw is discussed in this section.

\subsection{Advantages}

The proposed system addresses key limitations in existing approaches through region-specific YOLO fine-tuning for improved generalization across diverse Indian road conditions and end-to-end automation from real time detection to web based visualization. The lightweight interface with Base64-encoded image popups ensures accessibility for technical and non-technical users while providing actionable temporal, spatial, and severity data for evidence based road management.

The system supports smart governance initiatives by enabling budget allocation optimization, contractor performance evaluation, and damage severity-based repair prioritization. The public facing interface fosters transparency and citizen engagement while facilitating data driven decision making that optimizes public spending through real time evidence rather than periodic manual inspections.

\subsection{Challenges}

The implementation of the pothole detection system encountered several technical challenges that required systematic solutions:

\begin{itemize}
    \item \textbf{Model Selection and Compatibility:} Multiple YOLO versions (v4, v5, v7, v8) were evaluated for detection performance. YOLOv8 was selected based on superior metrics, necessitating adaptation of dependencies and inference scripts across version transitions.
    
    \item \textbf{Domain-Specific Dataset Requirements:} Generic YOLO models exhibited poor performance on Indian road conditions due to domain mismatch. Challenges included varied pothole characteristics (wet/dry, deep/shallow), region-specific visual noise (tar patches, oil marks), and diverse environmental conditions. A custom dataset comprising thousands of manually annotated dashcam frames was developed using the Roboflow platform and enhanced through data augmentation techniques.
    
    \item \textbf{Temporal Synchronization Issues:} 
    \begin{itemize}
        \item Initial YOLO based timestamp extraction failed due to model limitations in text recognition and variable overlay positioning
        \item OCR based extraction was implemented with region of interest identification
        \item Extracted timestamps suffered from formatting inconsistencies, delimiter misrecognition (":" as ".", "/" as "1"), and variable millisecond inclusion
        \item Multiple regex patterns with delimiter normalization were implemented for robust parsing
    \end{itemize}
    
    \item \textbf{GPS-Video Synchronization:} GPS timestamps (UTC) and dashcam overlays (IST/local time) required calibration with a measured offset of 5 hours, 30 minutes, 44 seconds. GPS coordinate precision was improved by rounding to 5 decimal places to mitigate signal jitter effects.
    
    \item \textbf{Spatial Redundancy Management:} High frequency processing (30fps) generated multiple detections per pothole, resulting in data redundancy and duplicate map markers. Haversine distance based deduplication with a 2.5 meter threshold using geopy.distance library effectively resolved spatial redundancy issues.
    
    \item \textbf{Image Storage and Retrieval Optimization:} Direct database storage of detection frames caused performance degradation due to size constraints and JSON API incompatibility. Base64 encoding was implemented to convert images to text format, enabling efficient database operations while maintaining frontend compatibility and eliminating file system dependencies.
\end{itemize}

\section{Conclusion and Future Work}

This paper presents a fully automated pothole detection and geotagging system combining YOLO based deep learning with OCR timestamp extraction and GPS synchronization for Indian road conditions. The system addresses critical limitations through spatial-temporal precision, metadata integration, and scalable solutions for frame redundancy and storage optimization, demonstrating reliable performance across diverse scenarios while supporting infrastructure planning and contractor performance auditing.

Future work will extend the system to comprehensive road quality assessment, including pavement distress classification, surface roughness estimation, and contractor quality monitoring through predictive degradation analysis. Advanced computer vision techniques such as semantic segmentation, 3D reconstruction, and multi-modal sensor fusion will enhance detection granularity, while integration with smart city infrastructure, including traffic management and IoT networks, will enable comprehensive urban mobility solutions and data driven governance in developing countries.


\begin{acks}
This work is partially funded by Netweb Technologies India Ltd. and RIN4001, Department of Atomic Energy. This work has benefited from the use of AI language tools (LLMs) for non-substantive editing, including phrasing and grammatical correction. All intellectual content is original and authored by the contributors.
\end{acks}

\bibliographystyle{ACM-Reference-Format}
\bibliography{sample-base}


\end{document}